\RequirePackage{amsmath}
\documentclass[runningheads]{llncs}

\usepackage{amsfonts,bm}
\usepackage{mathtools}
\usepackage{graphicx}
\usepackage{float}
\usepackage{hyperref} 
\usepackage{cuted}
\usepackage{lipsum, color}
\usepackage{latexsym,amssymb}
\usepackage{multirow}
\usepackage{url}

\usepackage{amsmath, amsthm,epsfig,bm,booktabs,amssymb}
\usepackage[ruled,vlined]{algorithm2e}
\usepackage{amsmath, amsfonts}
\usepackage{amsmath,latexsym,amssymb,epsf,amsfonts,amsthm}
\usepackage{epsf}
\usepackage{color}
\usepackage{mathrsfs}
\usepackage{epstopdf}
\usepackage{threeparttable}
\usepackage{graphicx}
\usepackage{tabularx}
\usepackage[all]{xy}

\newcommand{\w}{\mathbf{w}}

\def \R {{\mathbb R}}

\def \Q {{\mathbb Q}}

\newcommand{\lam}{\mbox{$\lambda$}}

\newcommand{\no}{\nonumber}

\newcommand{\be}{\begin{equation}}
\newcommand{\ee}{\end{equation}}
\newcommand{\ba}{\begin{eqnarray}}
\newcommand{\ea}{\end{eqnarray}}
\newcommand{\bi}{\begin{itemize}}
\newcommand{\ei}{\end{itemize}}

\newcommand{\real}{\mathbb{R}}

\newcommand{\comments}[1]{}

\begin{document}
\title{\texorpdfstring{A Channel-Pruned and Weight-Binarized\\ Convolutional Neural Network\\ for Keyword Spotting}{Lg}}
\titlerunning{Channel Pruned and Weight Binarized CNN}

\author{Jiancheng Lyu \inst{1} 
\and Spencer Sheen \inst{2}}

\institute{
UC Irvine, Irvine, CA 92697, USA.
\email{jianchel@uci.edu}
\and 
UC San Diego, La Jolla, CA 92093, USA.
\email{spsheen97@gmail.com.}
%Stanford, CA 94305, USA.
%\email{gsheen@stanford.edu}
}
\authorrunning{J. Lyu, S. Sheen}

\maketitle

\begin{abstract}
We study channel number reduction  
in combination with weight binarization (1-bit weight precision) to 
trim a convolutional neural network for a keyword spotting (classification) task. We adopt a group-wise splitting method based on the group Lasso penalty to
achieve over 50 \% channel sparsity while maintaining 
the network performance within 
0.25 \%  
accuracy loss. We show an effective three-stage procedure to balance 
accuracy and sparsity in network training. 

\end{abstract}
\bigskip

%\hspace{.12 in} {\bf AMS subject classifications:} 92B20, 65K10, 90C26.

%\newpage 

%\setcounter{page}{1}
\section{Introduction}
\setcounter{equation}{0}
Reducing complexity of neural networks while maintaining their performance is both fundamental and practical for resource limited platforms such as mobile phones. In this paper, we integrate two methods, namely channel pruning and weight quantization, to trim down  the number of parameters for a keyword spotting 
convolutional neural network (CNN, \cite{sp15}).
%and accelerate its inference.  
\medskip

Channel pruning aims to lower the number of 
convolutional channels, which is a group sparse 
optimization problem. Though group Lasso penalty 
\cite{GL_2006} is known in statistics, and has been 
applied directly in gradient decent training 
of CNNs \cite{GLDNN_2016} earlier, we found that 
the direct approach is not effective to realize sparsity for the keyword CNN \cite{sp15,tf18}. 
Instead, we adopt a group version of 
a recent relaxed variable splitting 
method \cite{DX_18}. This relaxed group-wise splitting 
method (RGSM, see \cite{rgsm} for the first study on deep image networks) accomplished over 50\% sparsity while keeping accuracy loss at a moderate level. 
In the next stage (II), the original network accuracy is recovered with a retraining of float precision weights while leaving out the pruned channels 
in stage I. In the last stage (III), the network weights are binarized into 1-bit precision 
with a warm start training based on stage II. 
At the end of stage III, a channel pruned (over 50 \%)
and weight binarized slim CNN is created with 
validation accuracy within 0.25 \% of that of 
the original CNN.

The rest of the paper is organized as follows. In section 2, we review the network architecture of keyword spotting CNN \cite{sp15,tf18}. In section 3, we introduce the proximal operator of group Lasso, RGSM, and its  
convergence theorem where an equilibrium condition is stated for the limit. We also 
outline binarization, the BinaryConnect (BC) training algorithm \cite{bc_15} 
and its blended version \cite{Yin18} 
to be used in our experiment. Through a 
comparison of BC and RGSM, we derive a hybrid 
algorithm (group sparse BC) which is of independent 
interest. In section 4, we describe our three stage training results, which indicate that RGSM is the most effective method and produces two slim CNN models 
for implementation.
%on a mobile app. 
Concluding remarks are in sections 4. 

\section{Network Architecture}
\begin{figure}[ht]
\begin{center}
\includegraphics[width=0.8\textwidth]{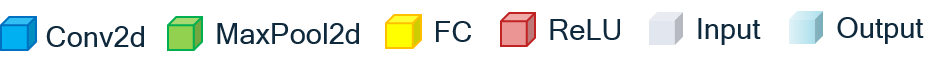}
\includegraphics[width=0.9\textwidth]{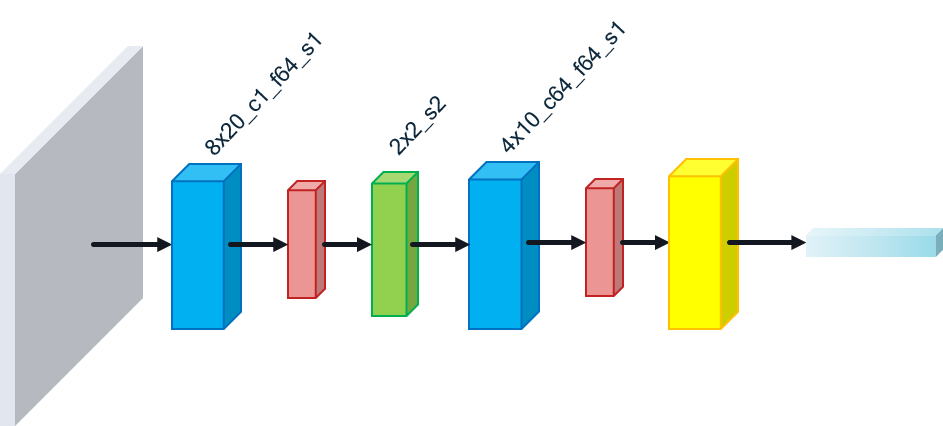}
\caption{A convolutional model with single-channel input. We use a simple notation to indicate conv. and max-pooling layers. For instance, 8x20\_c1\_f64\_s1 indicates a conv. layer with kernel size $8\times20$, channel number $1$, filter number $64$ and stride $1$.} \label{conv}
\end{center}
\end{figure}

Let us briefly describe the architecture of keyword CNN \cite{sp15,tf18} to 
classify a one second audio clip as either silence, an unknown word, `yes', `no', `up', `down', `left', `right', `on', `off', `stop', or `go'.
After pre-processing by windowed Fourier transform, the input becomes 
a single-channel image (a spectrogram) of size $t \times f$, 
same as a vector $v \in \R^{t\times f}$, where $t$ and $f$ are the input feature dimension 
in time and frequency respectively. Next is a convolution layer that operates as follows.
A weight tensor $W \in \R^{(m\times r)\times1\times n}$ is 
convolved with the input $v$. The weight tensor is a  
local time-frequency patch of size $m\times r$, where $m \leq t$ and $r \leq f$. 
The weight tensor has $n$ hidden units (feature maps), and may down-sample 
(stride) by a factor $s$ in time and $u$ in frequency. The output of the 
convolution layer is $n$ feature maps of size 
$(t - m + 1)/s \times (f-r +1)/u$. Afterward, a max-pooling operation replaces 
each $p\times q$ feature patch in time-frequency domain by the maximum value, which 
helps remove feature variability due to speaking styles, distortions etc. 
After pooling, we have $n$ feature maps of size 
$(t - m + 1)/(s\, p) \times (f-r +1)/(u\, q)$. 
An illustration is in Fig. \ref{conv}. 
The keyword CNN has two convolutional (conv) layers and a fully connected layer. There is 1 channel 
in the first conv. layer and there are 64 channels in the second. The weights in the second conv. layer form a 4-D tensor $W^{(2)}\in \real^{W\times H\times C\times N}$, where $(W, H, C, N)$ are dimensions of spatial width, spatial height, channels and filters, $C=64$.
 
\medskip

% \begin{figure}
% \begin{center}
% \includegraphics[width=1.00\textwidth]{convlayer.png}
% \caption{A convolutional layer performing convolution with stride, and max-pooling in a single channel.} \label{conv}
% \end{center}
% \end{figure}

\section{Complexity Reduction and Training Algorithms}

\subsection{Group Sparsity and Channel Pruning}
 Our first step is to trim the 64 channels in the second conv. layer to a smaller number while maintaining network performance. Let weights in each channel form a group, then this becomes a group sparsity problem for which group Lasso (GL) has been a classical differentiable penalty \cite{GL_2006}.
  Let vector 
  \[w = (w_1, \cdots, w_g, \cdots, w_G), \; w_g \in \R^d, \; w\in \R^{d\times G},\]
  where $G$ is the number of groups. Let $I_g$ be the indices of $w$ in group $g$. The group-Lasso penalty is \cite{GL_2006}:
\be
\|w\|_{GL}:= \sum_{g=1}^{G}\, \| w_g\|_{2}. \label{GL1}
\ee
 It is easy to implement GL as an additive penalty term for deep neural network training \cite{GLDNN_2016} by minimizing a penalized objective function of the form:
 \be
f(w):=\ell (w) + \mu \, P(w), \;\;\; 
\mu > 0, \label{ObjGL}
 \ee
where $\ell (w)$ is a standard loss function on data such as cross entropy \cite{YD_15}, and $P(w)$ is a penalty function equal to sum of weight decay ($\ell_2$ norm all network weights) and GL. For ease of notation, we merge the weight decay term with $\ell$ and take $P$ as GL below.
\medskip

In a case study of training CNN with un-structured weight sparsity \cite{Xue_19}, a direct minimization 
of $\ell_1$ type penalty 
as an additive term in the training objective 
function provides less sparsity and accuracy (Table 4 of \cite{Xue_19}) than 
the Relaxed Variable Splitting Method (RVSM \cite{DX_18}). In the group sparsity setting here, 
we shall see that the direct minimization of GL in (\ref{ObjGL}) is also not efficient.
Instead, we adopt a group version of 
RVSM \cite{DX_18}, which minimizes 
the following Lagrangian function of $(u,w)$ 
alternately:
  \be
    \mathcal{L}_{\beta}(u,w) = \ell (w) + \, \mu \, P(u) + \frac{\beta}{2}\lVert w-u \rVert^2_2, 
    \label{Lagr}
  \ee
for a parameter $\beta > 0$. 
\medskip

The $u$-minimization is 
in closed form for GL. To see this, 
consider finding the GL proximal (projection) operator by solving:
\be
y^{*} = {\rm argmin}_{y}\; \frac{1}{2}\, \| y - w \|^2 + \lam\, \| y \|_{GL}, \label{GL4}
\ee 
for parameter $\lam >0 $ or group-wise:
\be
y_{g}^{*}= {\rm argmin}_{y_g}\; \lam \, \|y_g\|+ 
\frac{1}{2}\, \sum_{i \in I_g} \| y_{g,i} - w_{g,i} \|^2.  \label{GL5}
\ee
If $y_{g}^{*} \not = 0$, the objective function of (\ref{GL5}) is differentiable and 
setting gradient to zero gives:
\[
y_{g,i}-w_{g,i} + \lam y_{g,i}/\|y_g\|  = 0,  \]
or:
\[ ( 1 + \lam /\|y_g\| ) y_{g,i} = w_{g,i}, \; \forall i \in I_g, \]
implying:
\[ ( 1 + \lam /\|y_g\| ) \, \|y_{g}\| = \|w_{g}\|, \]
or:
\be
\|y_{g}^{*}\| = \|w_{g}\| - \lam, \;\; {\rm if}\; \|w_g\| > \lam. \label{GL6}
\ee
Otherwise, the critical equation does not hold and $y_{g}^{*} = 0$.
The minimal point formula is:
\[
y_{g,i}^{*} = w_{g,i} ( 1+ \lam / (\|w_g\| -\lam))^{-1} = w_{g,i} (\|w_g\| - \lam)/\|w_g\|, \;{\rm if}\; \|w_g\| > \lam;
\]
%if $\|w_g\| > \lam$; 
otherwise, $y_{g}^{*} =0$. The result can be written as a 
soft-thresholding operation:
\be
y_{g}^{*}={\rm Prox}_{GL,\lam} (w_g) := w_{g} \, \max(\|w_g\| - \lam, 0)/\|w_g\| 
%&:=& w_g \, S_{\lam}(\|w_g\|)/\|w_g\|. 
\label{GL7}
\ee
\medskip

The $w$ minimization is by 
gradient descent, implemented in practice as 
stochastic gradient descent (SGD). Combining the $u$ and $w$ updates, we have the 
Relaxed Group-wise Splitting Method (RGSM):
\begin{eqnarray}
u_{g}^{t} &=& {\rm Prox}_{GL,\lambda}(w_{g}^{t}), \;\; 
g=1,\cdots, G, 
\no \\
w^{t+1}&= &w^t - \eta\, \nabla \ell (w^t)
-\eta\, \beta\, (w^t-u^t),
\label{GL9}
\end{eqnarray}
where $\eta$ is the learning rate.

\subsection{Theoretical Aspects}
The main theorem of \cite{DX_18} guarantees the convergence of RVSM algorithm under some conditions on the parameters $(\lambda,\beta,\eta)$ and initial weights in case of one convolution layer network and Gaussian input data. The latter conditions are used to prove that the loss function $\ell$ obeys Lipschitz gradient inequality on the iterations. Assuming that the Lipschitz gradient condition holds for $\ell$, we adapt the 
main result of \cite{DX_18} into: 
    \begin{theorem}\label{conv_thm}
    Suppose that $\ell$ is bounded from below, and satisfies the Lipschitz gradient inequality: $\|\nabla \ell(x)-\nabla \ell (y)\|\leq L\, \|x-y\|$, $\forall (x,y)$, 
    for some positive constant $L$. 
    Then there exists a positive constant $\eta_0 = \eta_0(L,\beta) \in (0,1)$ so that if $\eta < \eta_0$, the Lagrangian function  $\mathcal{L}_{\beta}(u^t,w^t)$ is descending and converging in $t$, with $(u^t,w^t)$ of RGSM algorithm satisfying $\|(u^{t+1},w^{t+1}) - (u^t,w^t)\| \to 0$ as $t \to +\infty$, and subsequentially approaching a limit point $(\bar{u}, \bar{w})$. The limit point $(\bar{u}, \bar{w})$ satisfies the {\it equilibrium} system of equations:
    \ba
\bar{u}_{g} & = & {\rm Prox}_{GL,\lambda}(\bar{w}_{g}), \; g=1,\cdots, G, \no  \\
\nabla \ell (\bar{w}) &=& \beta \, (\bar{u} - \bar{w}). \label{ste}
\ea
    \end{theorem}
\medskip

\begin{remark}
The system (\ref{ste}) 
serves as a ``critical point condition''.
The $\bar{u}$ is the desired weight vector 
with group sparsity that network training aims to 
reach.
\end{remark}

\begin{remark}
The group-$\ell_0$ penalty is: 
\be
\|w\|_{GL0}:= \sum_{g=1}^{G}\, 1_{(w_g: \| w_g\|_{2}\not = 0)} \label{GL2}
\ee
 Then the GL proximal problem (\ref{GL5}) is replaced by:
\be
y_{g}^{*}= {\rm argmin}_{y_g}\; \lam \, 1_{\|y_g\|\not = 0} + 
\frac{1}{2}\, \sum_{i \in I_g} \| y_{g,i} - w_{g,i} \|^2.  \label{GL8}
\ee
If $y_g = 0$, the objective equals $\|w_g\|_{2}^{2}/2$. So if 
$\lam \geq \|w_g\|^2/2$, $y_g = 0$ is a minimal point. 
If $\lam < \|w_g\|^2/2$, $y_g = w_g$ gives minimal value $\lam$. 
Hence the thresholding formula is:
\be 
y_{g}^{*} := {\rm Prox}_{GL0,\lam} (w_g) = w_g \; 1_{\|w_g\|_{2} > \sqrt{2\lam}}. \label{GL8b}
\ee
Theorem \ref{conv_thm} remains true with 
(\ref{ste}) modified where ${\rm Prox}_{GL,\lambda}$
is replaced by ${\rm Prox}_{GL0,\lam}$.
\end{remark}

\subsection{Weight Binarization}
The CNN computation can speed up a lot if the weights 
are in the binary vector form: float precision scalar times a sign vector $(\cdots, \pm 1,\pm 1, \cdots)$, see 
\cite{xnornet_16}. For the keyword CNN, such weight binarization alone doubles the speed of an Android 
app that runs on Samsung Galaxy J7 cellular phone \cite{SL_19} 
with standard tensorflow functions such as 
`conv2d' and `matmul'. 
\medskip

Weight binarized network training involves a 
projection operator or the solution of finding the closest binary vector to a given 
real vector $w$. The projection is written as 
${\rm proj}_{\Q} \, w$, for $w \in \R^D$, $\Q =\R_{+}\times \{\pm 1\}^D$. 
When the distance is Euclidean (in the sense of $\ell_2$ norm $\|\cdot \|$), the problem:
\be
{\rm proj}_{\Q,a} (w):= {\rm argmin}_{z \in \Q}\; \|z - w\| \label{p1}
\ee
has exact solution \cite{xnornet_16}: 
\be
{\rm proj}_{\Q,a} (w) = {\frac{\sum_{j=1}^{D} |w_j|}{D}} \; {\rm sgn}(w) \label{p2}
\ee
where ${\rm sgn}(w)=(q_1,\cdots,q_j,\cdots, q_D)$, and  \[
 q_{j} = \left \{ \begin{array}{rr}
              1  & {\rm if}\;  w_j \geq 0 \\
             -1 &  \; {\rm otherwise.}
          \end{array} \right.
\]
The projection is simply the sgn function of $w$ times the arithmetic average of the absolute values of the components of $w$. 
\medskip

The standard training algorithm for binarized weight network is BinaryConnect \cite{bc_15}:
\be
\w_f^{t+1} = \w_f^t - \eta \, \nabla \ell(\w^t), \; \w^{t+1} = {\rm proj}_{\Q,a}(\w_f^{t+1}), \label{alg1}
\ee
where $\{\w^t\}$ denotes the sequence of binarized weights, and $\{\w^t_f\}$ is an auxiliary sequence of floating weights (32 bit). Here we use the blended 
version \cite{Yin18}:
\be
\w_f^{t+1} =(1-\rho)\, \w_{f}^{t} + \rho \, \w^t  - \eta \, \nabla \ell(\w^t), \; \w^{t+1} = {\rm proj}_{\Q,a}(\w_f^{t+1}), \label{alg2}
\ee
for $0<\rho \ll 1 $. The algorithm (\ref{alg2}) becomes the  classical projected gradient descent at $\rho=1$, 
which suffers from weight stagnation 
due the discreteness of $\w^t$ however. The blending in (\ref{alg2}) leads to a better theoretical property \cite{Yin18} that the sufficient descent inequality holds if the loss function $\ell$ has Lipschitz gradient. 
\medskip

\begin{remark}
In view of (\ref{GL9}) and (\ref{alg2}), we see an interesting connection that both involve a projection step, as ${\rm Prox}$ is a projection in essence.
The difference is that $\nabla \ell$ in BC 
is evaluated at the projected weight $\w^t$. If we 
mimic such a BC-gradient, and evaluate the 
gradient of Lagrangian in $w$ at $u^t$ instead of $w^t$, then (\ref{GL9}) becomes:
\begin{eqnarray}
u_{g}^{t} &=& {\rm Prox}_{GL,\lambda}(w_{g}^{t}), \;\; 
g=1,\cdots, G, 
\no \\
w^{t+1}&= &w^t - \eta\, \nabla \ell (u^t).
\label{GL10}
\end{eqnarray}
We shall call (\ref{GL10}) a Group Sparsity BinaryConnect (GSBC) algorithm and compare it with RGSM in our experiment. 
\end{remark}

\section{Experimental Results}
In this section, we show training results of channel pruned and weight binarized audio CNN 
based on GL, RGSM, and GSBC. We assume that 
the objective function under gradient descent is $\ell (\cdot)+ \mu \, \|\cdot \|_{GL}$, with a threshold parameter $\lam$. For GL, $\mu>0$, $\lam=0$, $\beta=0$. For RGSM, $\mu=0$, $\lam > 0$, $\beta=1$. For GSBC, $\mu=0$, $\lam > 0$, $\beta = 0$. The experiment 
was conducted in TensorFlow on a single GPU machine with NVIDIA GeForce GTX 1080. The overall architecture \cite{tf18} consists of two convolutional layers, 
one fully-connected layer followed by a softmax function to output class probabilities. The training loss $\ell(\cdot)$ is the standard cross entropy function. The learning rate begins at $\eta =0.001$, and is reduced by a factor of 10 in the 
late training phase. The training 
proceeds in 3 stages:
\begin{itemize}
    \item Stage I: channel pruning with a suitable choice of $\mu$ or $\lam$ so that sparsity emerges at a moderate accuracy loss.
    \medskip
    
    \item Stage II: retrain float precision (32 bit) weights in the un-pruned channels at the fixed channel sparsity of Stage I, aiming to recover the lost accuracy in Stage I.
    \medskip
    
    \item Stage III: binarize the weights in each layer with warm start from the pruned network of Stage II, aiming to nearly maintain the accuracy in Stage II.
\end{itemize}

\begin{figure}[ht!]
      \centering
       \includegraphics[width=0.8\textwidth]{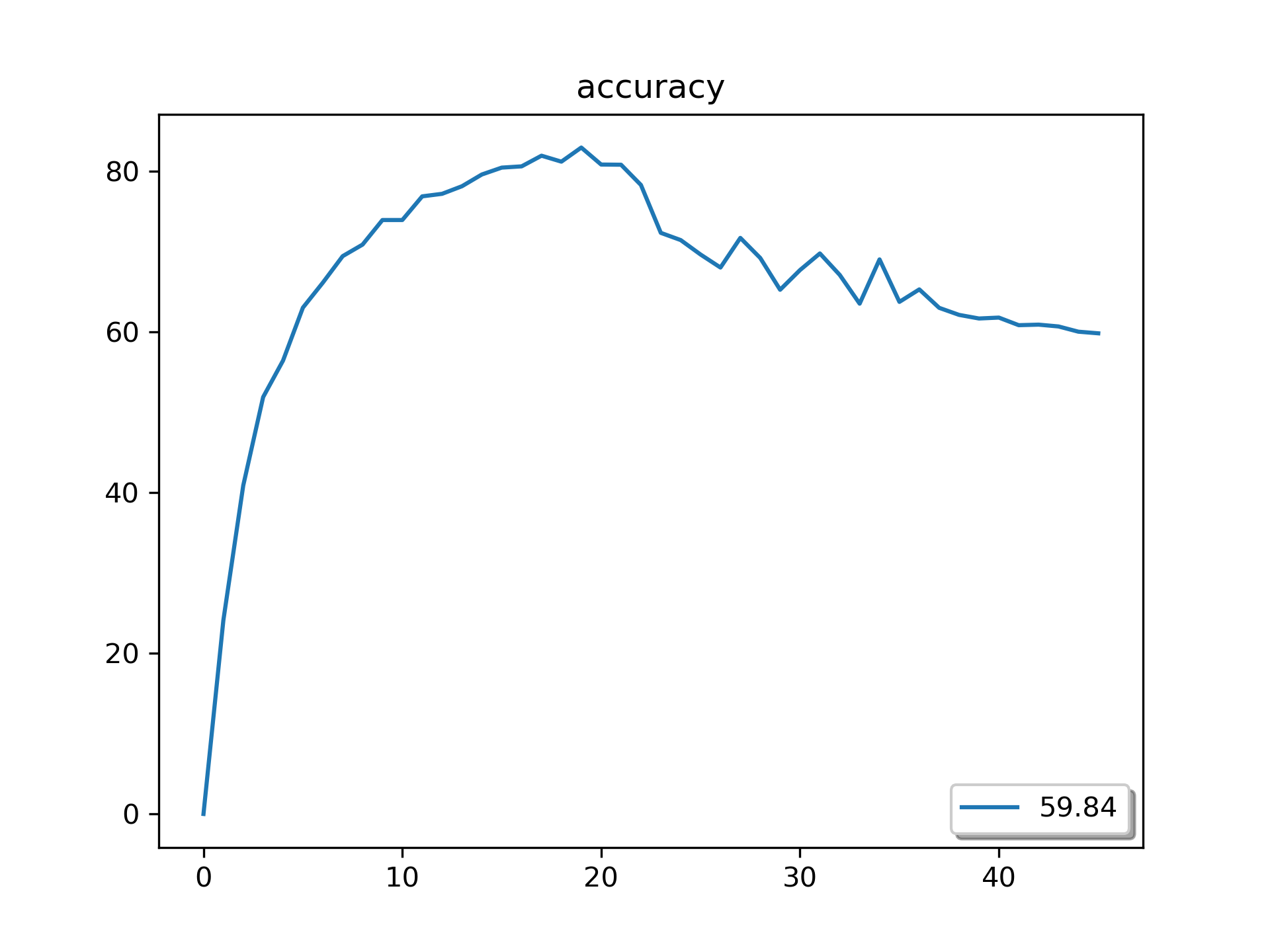}
     \caption{Validation accuracy vs. number of epochs in Stage-1 training by RGSM at $(\lam,\beta)=(0.05,1)$. The accuracy at the last epoch is 59.84 \%.}
       \label{chan_prun1}
 \end{figure}

\begin{figure}[ht!]
      \centering
       \includegraphics[width=0.8\textwidth]{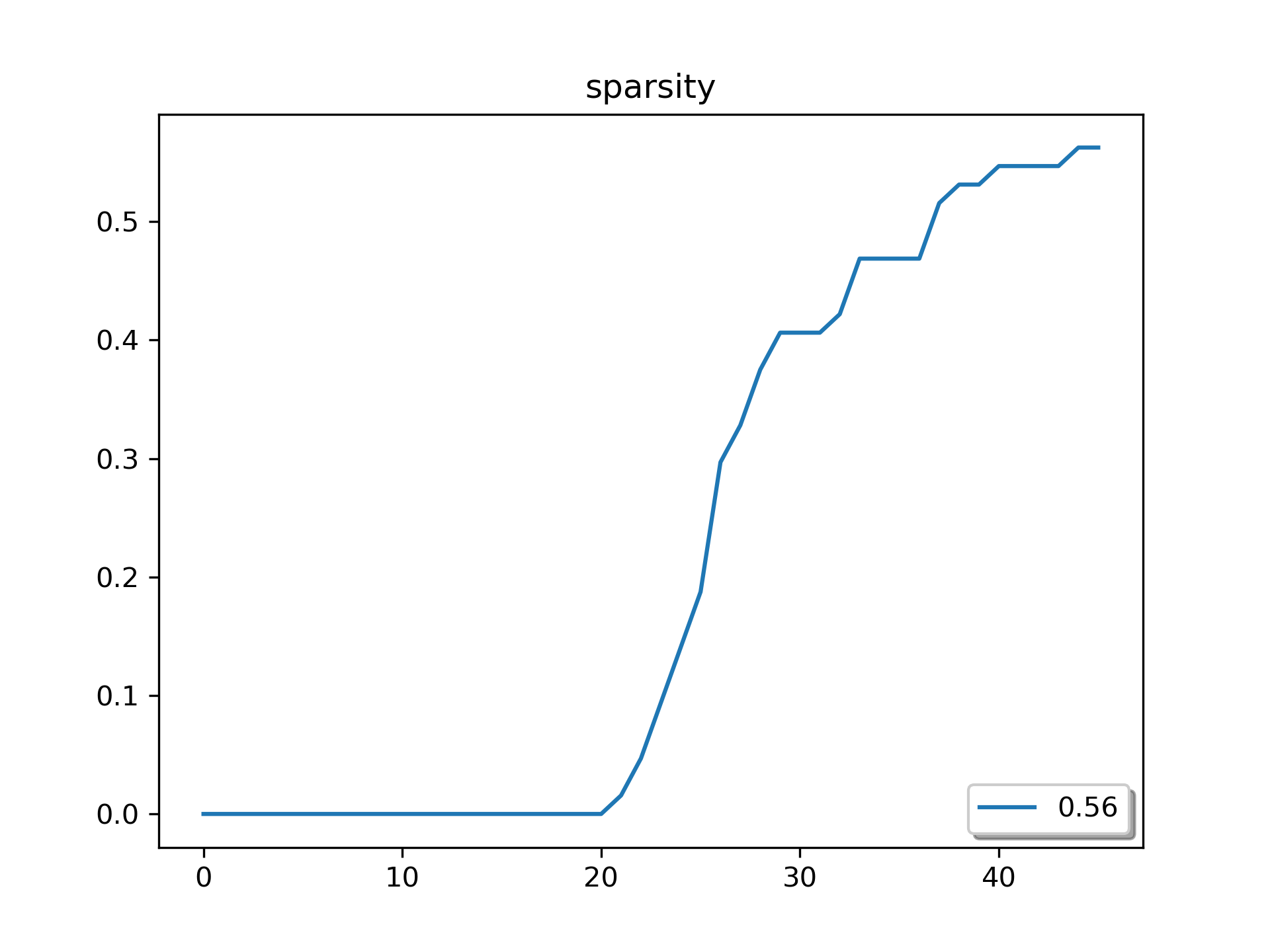}
     \caption{Channel sparsity vs. number of epochs in Stage-1 training by RGSM at $(\lam,\beta)=(0.05,1)$. The sparsity at the last epoch is  56.3 \%.}
       \label{chan_prun2}
 \end{figure}

\begin{figure}[ht!]
      \centering
       \includegraphics[width=0.9\textwidth]{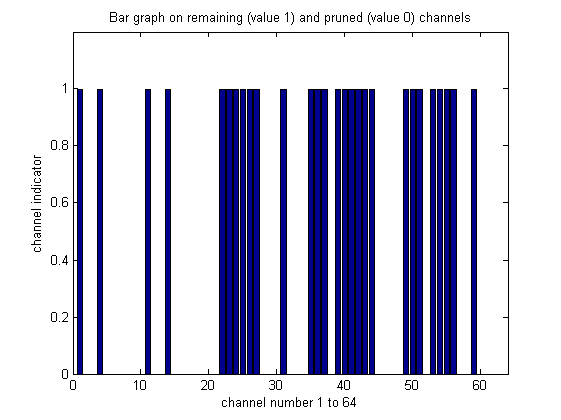}
     \caption{Remaining channels illustrated by bars vs. channel number (1 to 64) after Stage-1 training by RGSM at $(\lam,\beta)=(0.05,1)$. The sparsity (\% of 0's) is 56.3 \%.}
       \label{chan_prun_bar}
 \end{figure}

Stage I begins with random (cold) start and performs 18000 iterations (default, about 50 epochs). Fig. \ref{chan_prun1} shows the validation accuracy of 
RGSM at $(\lam,\beta)=(0.05,1)$ vs. epoch number. 
The accuracy climbs to a peak value above 80 \% at epoch 20, then comes down and ends at 59.84 \%. 
The accuracy slide agrees with channel sparsity gain 
beginning at epoch 20 and steadily increasing 
to nearly 56 \% at the last epoch seen in Fig. \ref{chan_prun2}. The bar graph in Fig. \ref{chan_prun_bar} shows the pruning pattern and the remaining channels (bars of unit height). 
At $(\lam,\beta)=(0.04,1)$, RGSM stage I training 
yields a higher validation accuracy 76.6 \% with a slightly lower channel sparsity 51.6 \%. At the same $(\lam,\beta)$ values, GSBC gives an even higher validation accuracy 80.9 \% but much lower channel sparsity of 26.6 \%. The GL method
produces minimal channel sparsity in the range 
$\mu \in (0,1)$ covering the 
corresponding $\lam $ value where sparsity emerges in RGSM. The reason appears to be that the network has certain internal constraints that prevent the GL penalty from getting too small. Our experiments show that even with the cross-entropy loss $\ell(\cdot)$ removed from the 
training objective, the GL penalty 
cannot be minimized below some positive level. The Stage-I results are tabulated in Table 
\ref{tab1} with a GL case at $\mu =0.6$. It is clear that RGSM is the best method 
to go forward with to stage II. 
\medskip

\begin{table}[!b]
\caption{Validation Accuracy (\%) and Channel (Ch.) Sparsity (\%) after Stage I (Ch. pruning).}
\label{tab1}
\begin{center}
\begin{tabular}{|c|c|c|c|c|c|}
\hline
\textbf{Model} & $\boldsymbol{\beta}$ & $\boldsymbol{\lambda}$ & $\boldsymbol{\mu}$ & \textbf{\textit{Accuracy}}& \textbf{\textit{Ch. Sparsity}} \\
\hline
 Original Audio-CNN & 0 & 0 & 0 & 88.5 & 0 \\
 \hline
 GL  Ch-pruning  & 0 & 0 & 0.6 & 66.8 & 0  \\
\hline
RGSM Ch-pruning & 1 & 4.e-2 & 0 & 76.6 & 51.6    \\
\hline
RGSM Ch-pruning & 1 & 5.e-2 & 0 & 59.8 & 56.3    \\
\hline
GSBC Ch-pruning & 0 & 4.e-2 & 0 & 80.9 &  26.6\\
\hline
%Ch-pruning by RGSM+GL & 1 & 3.8e-2 & 1.e-4 & 76.0 &  48.0\\
%\hline
\end{tabular}
\end{center}
\end{table}

\begin{figure}[ht]
\begin{center}
\includegraphics[width=0.9\textwidth]{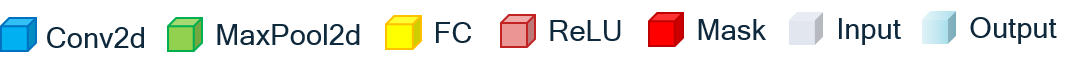}
\includegraphics[width=1.0\textwidth]{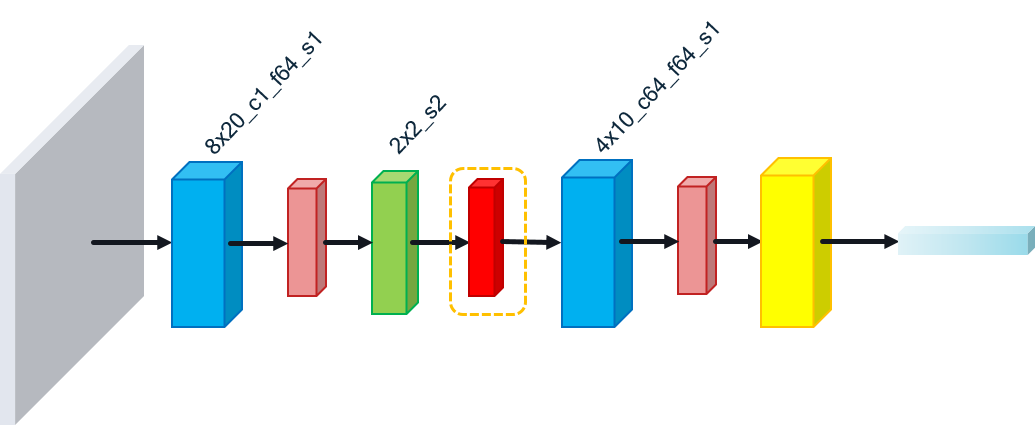}
\caption{The CNN model with pruned channels masked out (the masking layer in red).} \label{conv_mask}
\end{center}
\end{figure}

\begin{figure}[ht!]
      \centering
       \includegraphics[width=0.9\textwidth]{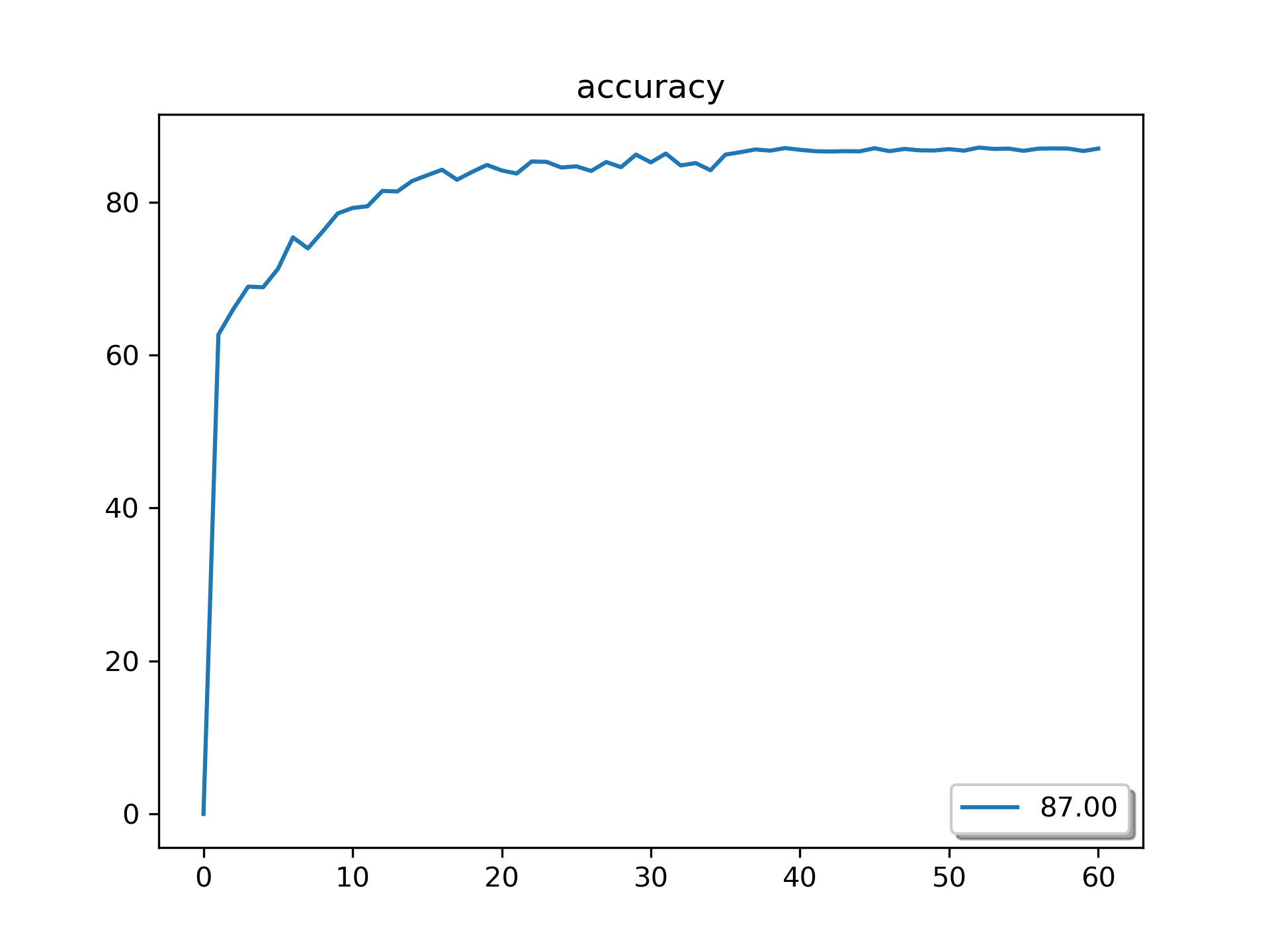}
     \caption{Validation accuracy vs. number of epochs in Stage-3 binary (1-bit) weight training. The accuracy at the 
     last epoch is $87\, \%$. Channel sparsity is 56.3 \%.}
       \label{chan_prun3}
 \end{figure}

In Stage II, we mask out the pruned channels to keep sparsity invariant (Fig. \ref{conv_mask}), and retrain float precision weights in the complementary part of the network. Fig. \ref{chan_prun3} shows that with a dozen epochs of retraining, the accuracy of the RGSM pruned model at $\lam=0.04$ ($0.05$) in Stage I reaches 89.2 \% (87.9 \%), at the level of the original audio CNN. 
\medskip

\begin{figure}[ht!]
      \centering
       \includegraphics[width=0.8\textwidth]{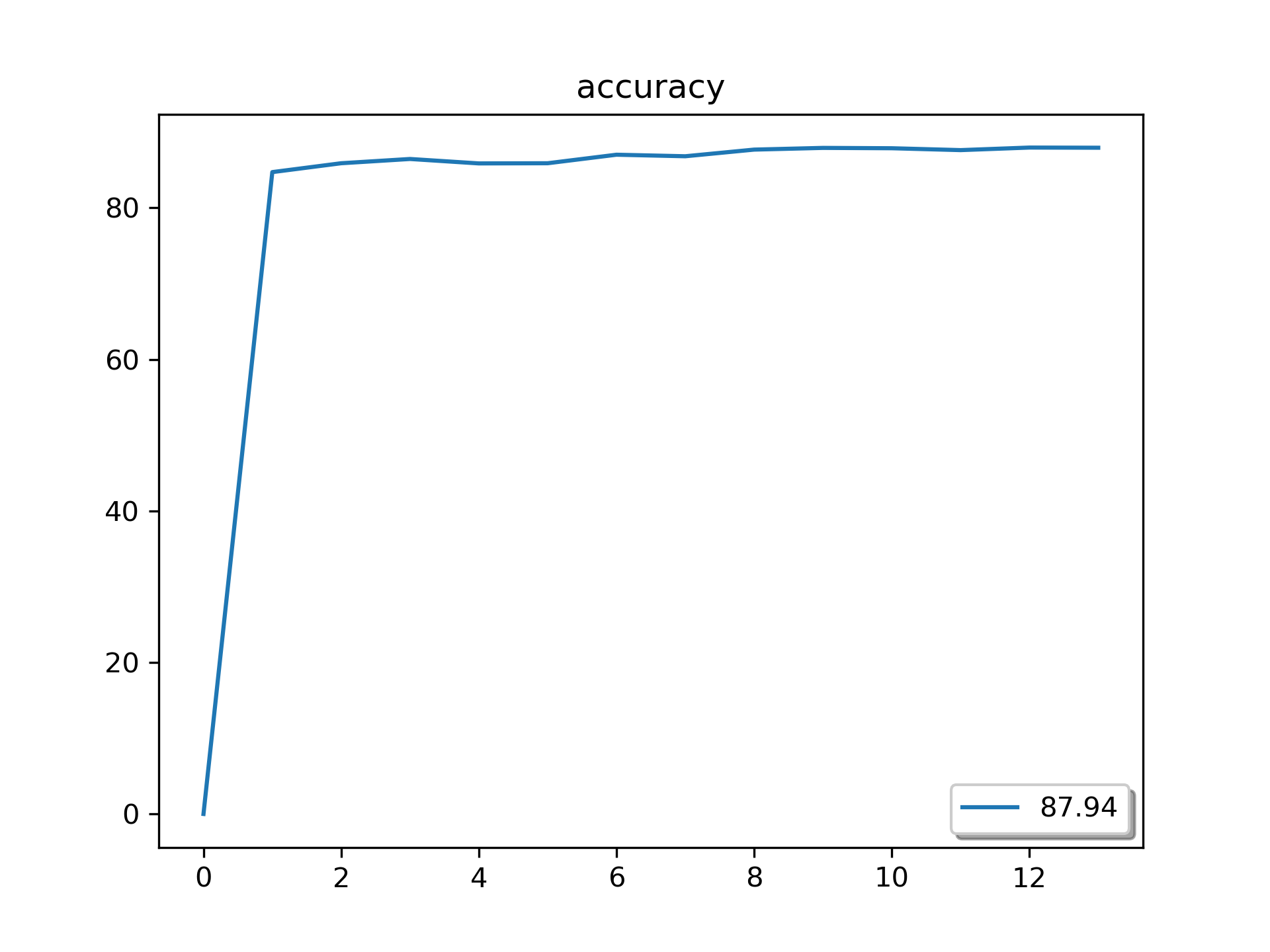}
     \caption{Validation accuracy vs. number of epochs in Stage-2 float (32 bit) weight retraining. The accuracy at the 
     last epoch is 87.94 \%. Channel sparsity is 56.3 \%. }
       \label{chan_prun4}
 \end{figure}

In Stage III, with blending parameter $\rho=$ 1.e -5, the weights in the network modulo the 
masked channels are binarized with validation accuracy 
88.3 \% at channel sparsity 51.6 \%, and 
87 \% at channel sparsity 56.3 \%, see Fig. \ref{chan_prun4} and Table \ref{tab3}.

\begin{table}[!b]
\caption{Validation Accuracy (\%) and Channel (Ch.) Sparsity (\%) after Stage II (float precision weight retraining).}
\label{tab2}
\begin{center}
\begin{tabular}{|c|c|c|c|c|c|}
\hline
\textbf{Model} & $\boldsymbol{\beta}$ & $\boldsymbol{\lambda}$ & $\boldsymbol{\mu}$ & \textbf{\textit{Accuracy}}& \textbf{\textit{Ch. Sparsity}} \\
\hline
 Original Audio-CNN & 0 & 0 & 0 & 88.5 & 0 \\
 \hline
RGSM Ch-pruning + Float Weight Retrain & 1 & 4.e-2 & 0 & 89.2 & 51.6    \\
\hline
RGSM Ch-pruning + Float Weight Retrain & 1 & 5.e-2 & 0 & 87.9 & 56.3    \\
\hline

\end{tabular}
\end{center}
\end{table}

\begin{table}[!b]
\caption{Validation Accuracy (\%) and Channel (Ch.) Sparsity (\%) after Stage III (weight binarization training).}
\label{tab3}
\begin{center}
\begin{tabular}{|c|c|c|c|c|c|}
\hline
\textbf{Model} & $\boldsymbol{\beta}$ & $\boldsymbol{\lambda}$ & $\boldsymbol{\mu}$ & \textbf{\textit{Accuracy}}& \textbf{\textit{Ch. Sparsity}} \\
\hline
 Original Audio-CNN & 0 & 0 & 0 & 88.5 & 0 \\
 \hline
RGSM Ch-pruning + Weight Binarization & 1 & 4.e-2 & 0 & 88.3 & 51.6    \\
\hline
RGSM Ch-pruning + Weight Binarization & 1 & 5.e-2 & 0 & 87.0 & 56.3    \\
\hline

\end{tabular}
\end{center}
\end{table}

\section{Conclusion and Future Work}
We successfully integrated a group-wise splitting 
method (RGSM) for channel pruning, float weight retraining and weight binarization to arrive at  
a slim yet almost equally performing CNN for keyword spotting. Since channel pruning involves architecture change, 
there is additional work 
to speed up 
a hardware implementation. Preliminary test on a MacBook Air with a CPU version of 
Tensorflow shows 
as much as 28.87 \% speed up by 
the network structure with float precision weight in  
Fig. \ref{conv_mask}. An efficient way to implement the masking layer without resorting to an element-wise tensor multiplication (especially on a mobile phone) is worthwhile for our future work. 

\medskip

We also plan to study other penalties \cite{DX_18} such as group-$\ell_0$ (transformed-$\ell_1$) in the 
RGSM framework as outlined in Remark 2, and extend the three stage process developed here to multi-level complexity reduction on larger CNNs and other applications in the future.

\section{Acknowledgements}
The work was supported in part by NSF grants IIS-1632935 and 
DMS-1854434 at UC Irvine.

%\bibliographystyle{plain}
%\bibliography{main}

\begin{thebibliography}{99}
\bibitem{bc_15}M. Courbariaux, Y. Bengio and J. David,
{\em BinaryConnect: Training Deep Neural Networks with Binary Weights during Propagations},
Conferencie on Neural Information Processing Systems (NIPS), pp. 3123-3131, 2015.
\medskip

\bibitem{DX_18}T. Dinh, J. Xin, ``Convergence of a relaxed variable splitting method for learning sparse neural networks via $\ell_1$, $\ell_0$, and transformed-$\ell_1$ penalties'', arXiv preprint arXiv:1812.05719. \medskip

%\bibitem{Welling} Louizos, C., Welling, M., Kingma, D.: Learning sparse neural networks through $\ell_0$ regularization. In:ICLR (2018). arXiv: 1712.01312v2 \medskip

\bibitem{xnornet_16} M. Rastegari, V. Ordonez, J. Redmon and A. Farhadi,
{\em XNOR-Net: ImageNet Classification Using Binary
Convolutional Neural Networks}, European Conference on Computer Vision (ECCV), 2016.
\medskip

\bibitem{sp15}T. Sainath and C. Parada, {\em Convolutional Neural Networks for Small-footprint Keyword Spotting}, 
Interspeech 2015, pp. 1478-1482, Dresden, Germany, Sept. 6-10.
\medskip

\bibitem{SL_19}S. Sheen, J. Lyu,
{\em Median Binary-Connect Method and A Binary Weight Convolutional Neural Network for Word Recognition}, arXiv:1811.02784; IEEE Data Compression Conference (DCC), 2019; DOI: 10.1109/DCC.2019.00116.
\medskip

\bibitem{tf18}Simple audio recognition tutorial, tensorflow.org, last access Aug. 10, 2019.
\medskip

\bibitem{GLDNN_2016}W. Wen, C. Wu, Y. Wang, Y. Chen, and H. Li, ``Learning structured sparsity in deep neural networks,'' in NIPS, 2016.
\medskip

\bibitem{GL_2006} M. Yuan and Y. Lin, ``Model selection and estimation in regression with grouped variables,'' Journal of the Royal Statistical Society, Series B, 68(1):49–-67, 2007.
\medskip

\bibitem{Xue_19}F. Xue, J. Xin,
``Learning Sparse Neural Networks via L0 and TL1 by a Relaxed Variable Splitting Method with Application to Multi-scale Curve Classification,'' 
arXiv preprint arXiv: 1902.07419; 
in Proc. World Congress Global Optimization, Metz, France, July, 2019.  
DOI:10.1007/978-3-030-21803-4\textunderscore80. 
%In: Le Thi H., Minh Le H., Pham Dinh T. (eds), Optimization of Complex Systems: Theory, Models, Algorithms and Applications, pp. 800-809, Advances in Intelligent Systems and Computing, v. 991, Springer, 2020. 
\medskip

\bibitem{rgsm} B. Yang, J. Lyu, S. Zhang, Y-Y Qi, J. Xin
``Channel Pruning for Deep Neural Networks via a
Relaxed Group-wise Splitting Method'', In Proc. of 2nd International  Conference on AI for Industries, Laguna Hills, CA,
Sept. 25-27, 2019


\bibitem{Yin18}P. Yin, S. Zhang, J. Lyu, S. Osher, Y-Y. Qi, J. Xin, ``Blended coarse gradient descent for full quantization of deep neural networks''. Research in the Mathematical Sciences \textbf{6}(1), 14 (2019). DOI:10.1007/s40687-018-0177-6. arXiv: 1808.05240. 
\medskip

%\bibitem{BR_18} P. Yin, S. Zhang, J. Lyu, S. Osher, Y. Qi and J. Xin, {\em BinaryRelax: A Relaxation Approach for Training Deep Neural Networks with Quantized Weights},arXiv preprint arXiv:1801.06313, 2018; SIAM Journal on Imaging Sciences, to appear.

\bibitem{YD_15}D. Yu, L. Deng: 
Automatic speech recognition: a deep learning approach. Signals and Communication Technology. Springer, New York (2015)

%\bibitem{ZX}Zhang, S., Xin, J.: Minimization of transformed $l_1 $ penalty: closed form representation and iterative thresholding algorithms. Comm. Math Sci. \textbf{15}(2), 511--537 (2017)


\end{thebibliography}

%\clearpage

\end{document}